\documentclass{llncs}
\usepackage{amsmath,graphicx,subfigure,amssymb}
\usepackage{algorithmic, algorithm,color}
\input{psfig.sty}

\usepackage{fancyheadings}
\begin{document}

\mainmatter

\title{MinMax Radon Barcodes \\for Medical Image Retrieval}
\author{H.R. Tizhoosh\inst{1}, Shujin Zhu\inst{2}, Hanson Lo\inst{3}, Varun Chaudhari\inst{4}, Tahmid Mehdi\inst{5}}
\institute{KIMIA Lab,University of Waterloo,Canada \email{tizhoosh@uwaterloo.ca}\\
\and School of Electronic \& Optical Eng., Nanjing Univ. of Sci. \& Tech., Jiangsu, China\\
\and Cheriton School of Computer Science, University of Waterloo, Canada\\
\and Department of Mathematics, University of Waterloo, Canada\\
\and Centre for Computational Mathematics in Industry and Commerce, University of Waterloo, Canada
}


\maketitle

\begin{abstract}
Content-based medical image retrieval can support diagnostic decisions by clinical experts. Examining similar images may provide clues to the expert to remove uncertainties in his/her final diagnosis. Beyond conventional feature descriptors, binary features in different ways have been recently proposed to encode the image content. A recent proposal is ``Radon barcodes'' that employ binarized Radon projections to tag/annotate medical images with content-based binary vectors, called barcodes. In this paper, MinMax Radon barcodes are introduced which are superior to ``local thresholding'' scheme suggested in the literature. Using IRMA dataset with 14,410 x-ray images from 193 different classes, the advantage of using MinMax Radon barcodes over \emph{thresholded} Radon barcodes are demonstrated. The retrieval error for direct search drops by more than 15\%. As well, SURF, as a well-established non-binary approach, and BRISK, as a recent binary method are examined to compare their results with MinMax Radon barcodes when retrieving images from IRMA dataset. The results demonstrate that MinMax Radon barcodes are faster and more accurate when applied on IRMA images. 
\end{abstract}

\section{Introduction}
Searching for similar images in archives with millions of digital images is a difficult task that may be useful in many application domains. We usually search for images via ``text'' (or meta-data). In such cases, which appear to be the dominant mode of image retrieval in practice, all images have been tagged or annotated with some textual descriptions. Hence, the user can provide his/her own search terms such as ``birds'', `red car'', or ``tall building campus'' to find images attached to these keywords. Of course, text-based image search has a very limited scope. We cannot annotate all images with proper keywords that fully describe the image content. This is sometimes due to the sheer amount of manpower required to annotate a large number of images. But, more importantly, most of the time it is simply not possible to describe the content of the image with words such that there is enough discrimination between images of different categories. For example, search for a ``breast ultrasound tumor'' may be relatively easy even with existing text-based technologies. However, looking for a ``lesion which is taller than wide and is highly spiculated'' may prove to be very challenging. Apparently, domains such as medical image analysis are not profiting much from the text-based image search.

Content-based image retrieval (CBIR) has been an active research field for more than two decades. CBIR algorithms are primarily trimmed toward describing the content of the image with non-textual attributes, for instance with some type of features. If we manage to extract good features from the image, then image search becomes a classification and matching problem that works based on visual clues and not based on the text. Under \emph{good} features we usually understand such attributes that are invariant to scale, translation, rotation, and maybe even some types of deformation. In other words, features are good if they can uniquely characterize each image category with respect to their what they contain (shape, colors, edges, textures, segments etc.). 

The literature on feature extraction is rich and vast. Methods like SIFT and SURF have been successfully applied to many problems. In more recent literature, we observe a shift from traditional feature descriptors to \emph{binary} descriptors. This shift has been mainly motivated by the tremendous increase in the size of image archives we are dealing with. Binary descriptors are compact with inherent efficiency for searching, properties that lend themselves nicely to deal with big image data. 

In this paper, we focus on one of the recently introduced binary descriptors, namely Radon barcodes (section \ref{sec:BKG}). We introduce a new encoding scheme for Radon barcodes to binarize the projections (section \ref{sec:minmax}). We employ the IRMA dataset with 14,410 x-ray images with 193 classes to validate the performance of the proposed approach (section \ref{sec:exp}). In order to complete the experimentations, two other established methods, namely SURF and BRISK, are for the first time tested on IRMA dataset as well to draw some more general conclusions with respect to the performance of the proposed MinMax Radon barcodes. 

\section{Background}
\label{sec:BKG}

The literature on CBIR in general, and on medical CBIR, in particular, is quite vast. Ghosh et al. \cite{ghosh2011} review online solutions for content-based medical image retrieval such as GoldMiner, FigureSearch, BioText, Yottalook, IRMA, Yale Image Finder and iMedline. Multiple surveys are available that review recent literature \cite{rajam2013}, \cite{dharani2013}. To recent approaches that have used IRMA dataset (see section \ref{subsec:IRMA}) belong  autoencoders for image area reduction \cite{camlica2015autoencoding} and local binary patterns (LBPs) \cite{ojala2002},\cite{ahonen2006},\cite{camlica2015medical}. 

Although binary images (or embeddings) have been used to facilitate image retrieval in different ways \cite{tizhoosh2012computer},\cite{tizhoosh2012method},\cite{Daugman2004},\cite{Arvancheh2006}, it seems that binarizing Radon projections to use them directly for CBIR tasks is a rather recent idea \cite{Tizhoosh2015}. Capturing a 3D object is generally the main motivation for Radon transform \cite{Radon1917}. There are many applications of Radon transform reported in literature \cite{zhao2013},\cite{hoang2012},\cite{jadhav2009}. Chen and Chen \cite{chen2008} introduced Radon composite features (RCFs) that transform binary shapes into 1D representations for feature calculation. Tabbone et al. \cite{tabbone2008} propose a histogram of the Radon transform (HRT) invariant to geometrical transformations. Dara et al. \cite{daras2006} generalized Radon transform to radial and spherical integration to search for 3D models of diverse shapes. Trace transform is also a generalization of Radon transform \cite{kadyrov2001} for invariant features via tracing lines applied on shapes with complex texture on a uniform background for change detection. 

SURF (Speeded Up Robust Features) \cite{bay2008speeded} is one of the most commonly used keypoint detectors and feature descriptors for various applications. BRISK (Binary Robust Invariant Scalable Keypoints) \cite{Leutenegger2011}, in contrast, is one of the recently introduced binary feature descriptors that appears to be one of the robust binary schemes for CBIR \cite{Choi2014}. We use both SURF and BRISK in our experiments for comparative purposes. For the first time, we report the accuracy of these methods on IRMA dataset \cite{Lehmann2003},\cite{Mueller2010}.

\section{Radon barcodes}
\label{sec:RBC}

\begin{figure}[tb]
\begin{center}
\includegraphics[width=0.50\columnwidth]{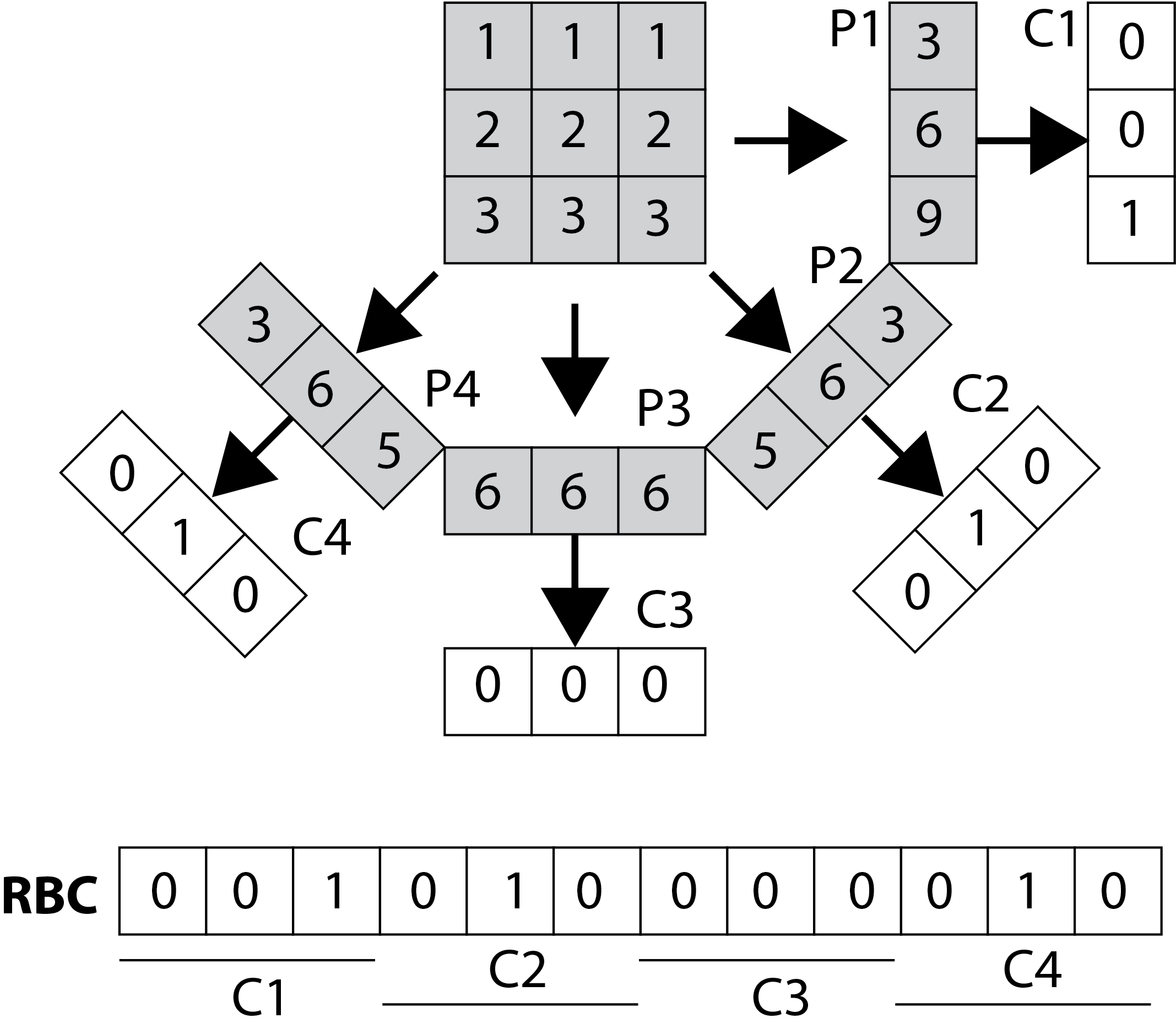}
\caption{Radon barcode (RBC) according to \cite{Tizhoosh2015} -- Parallel projections (here P1 to  P4) are binarized to create code fragments C1 to C4. Putting all code fragments together delivers the barcode \textbf{RBC}. }
\label{fig:RBC}
\end{center}
\end{figure}

The idea of Radon barcodes was introduced recently \cite{Tizhoosh2015},\cite{Tizhoosh2016}. Examining an image $I$ as a 2D function $f(x,y)$, one can project $f(x,y)$ along a number of parallel projection directions $\theta$. A projection is the sum (integral) of $f(x,y)$ values along lines constituted by each angle $\theta$ to create a new image $R(\rho,\theta)$ with $\rho = x \cos \theta + y \sin \theta$. Hence, using the Dirac delta function $\delta(\cdot)$ the Radon transform can be given as 
\begin{equation}
R(\rho,\theta) = \int\limits_{-\infty}^{+\infty} \int\limits_{-\infty}^{+\infty} f(x,y) \delta(\rho-x\cos \theta-y\sin\theta) dx dy.
\end{equation}
If we binarize all projections (lines) for individual directions using a ``local'' threshold for that angle (as proposed in \cite{Tizhoosh2015}), then we can assemble a barcode of all binarized projections as depicted in Figure \ref{fig:RBC}. A straightforward method to binarize the projections is to set a representative (or typical) value. This can be done by calculating the median value of all non-zero projection values as initially proposed in \cite{Tizhoosh2015}. 

Algorithm \ref{alg:Radon} describes the generation of \textbf{Radon barcodes (RBC)}  \footnote{Matlab code taken from http://tizhoosh.uwaterloo.ca/}. In order to receive same-length barcodes \emph{Normalize$(I)$} resizes all images into $R_N\times C_N$ images (i.e. $R_N= C_N=2^n,n\in \mathbb{N}^+$).

\begin{algorithm}[htb]
\caption{Radon Barcode (RBC) Generation \cite{Tizhoosh2015}}
\begin{algorithmic}[1]
\label{alg:Radon}
\STATE Initialize Radon Barcode $\mathbf{r} \leftarrow \emptyset$ 
\STATE Initialize: angle $\theta \leftarrow 0$, $\theta_{\max} = 180$, and image size $R_N=C_N\leftarrow 32$
\STATE $\bar{I} = \textrm{Normalize}(I,R_N,C_N)$ 
\STATE Set the number of projection angles, e.g. $n_p \leftarrow 8$
\WHILE{$\theta < \theta_{\max}$}
\STATE Get all projections $\mathbf{p}$ for $\theta$
\STATE Find typical value $T_\textrm{typical}\leftarrow\textrm{median}_i (\mathbf{p}_i)|_{\mathbf{p}_i \neq 0}$
\STATE Binarize projections: $\mathbf{b} \leftarrow \mathbf{p} \geq T_\textrm{typical}$ 
\STATE Append the new row $\mathbf{r} \leftarrow \textrm{append}(\mathbf{r},\mathbf{b} )$ 
\STATE $\theta \leftarrow \theta + \frac{\theta_{\max}}{n_p}$
\ENDWHILE
\STATE Return $\mathbf{r}$
\end{algorithmic}
\end{algorithm}

\section{MinMax Radon Barcodes}
\label{sec:minmax}

The thresholding method introduced in \cite{Tizhoosh2015} to binarize Radon projections is quite simple, hence, it may lose a lot of information that could contribute to the uniqueness of the barcode. For instance, employing a local threshold will not capture the general curvature of the projections. In contrast, if we examine how the projection values transit between local extrema, this may provide more expressive clues for capturing the shape characteristics of the scene/image depicted in that specific angle. 

Algorithm \ref{alg:minmaxRadon} provides the general steps for generating MinMax Radon barcodes. The smoothing function (Algorithm \ref{alg:minmaxRadon}, line 7) just applies a moving average to remove small peaks/valleys. We then can detect all peaks (maximums) and valleys (minimums) (Algorithm \ref{alg:minmaxRadon}, line 8). Subsequently, we locate all values that are on the way to transit from min/max to max/min, respectively (Algorithm \ref{alg:minmaxRadon}, lines 9-10). The projection can then be encoded by assigning corresponding values of zeros or ones (Algorithm \ref{alg:minmaxRadon}, lines 11-13). These are the main differences to the Radon barcode (Algorithm \ref{alg:Radon}).

\begin{algorithm}[htb]
\caption{MinMax Radon Barcodes}
\begin{algorithmic}[1]
\label{alg:minmaxRadon}
\STATE Initialize Radon Barcode $\mathbf{r} \leftarrow \emptyset$ 
\STATE Initialize: angle $\theta \leftarrow 0$, $\theta_{\max} = 180$, and image size $R_N=C_N\leftarrow 32$
\STATE $\bar{I} = \textrm{Normalize}(I,R_N,C_N)$ 
\STATE Set the number of projection angles, e.g., $n_p \leftarrow 8$
\WHILE{$\theta < \theta_{\max}$}
\STATE Get all projections $\mathbf{p}$ for $\theta$
\STATE Smooth $\mathbf{p}$: $\mathbf{\bar{p}} \leftarrow$ Smooth($\mathbf{p}$)
\STATE Find all minimums and maximums of $\mathbf{\bar{p}}$
\STATE $b_{\min}\leftarrow$ Find all $\mathbf{\bar{p}}$ bins that are in a min-max interval
\STATE $b_{\max}\leftarrow$ Find all $\mathbf{\bar{p}}$ bins that are in a max-min interval 
\STATE $\mathbf{b} \leftarrow \mathbf{\bar{p}}$
\STATE Set bits: $\mathbf{b}(b_{\min}) \leftarrow 0$; $\mathbf{b}(b_{\max}) \leftarrow 1$
\STATE Append the new row $\mathbf{r} \leftarrow \textrm{append}(\mathbf{r},\mathbf{b} )$ 
\STATE $\theta \leftarrow \theta + \frac{\theta_{\max}}{n_p}$
\ENDWHILE
\STATE Return $\mathbf{r}$
\end{algorithmic}
\end{algorithm}

\begin{figure}[tb]
\begin{center}
\includegraphics[width=0.7\columnwidth]{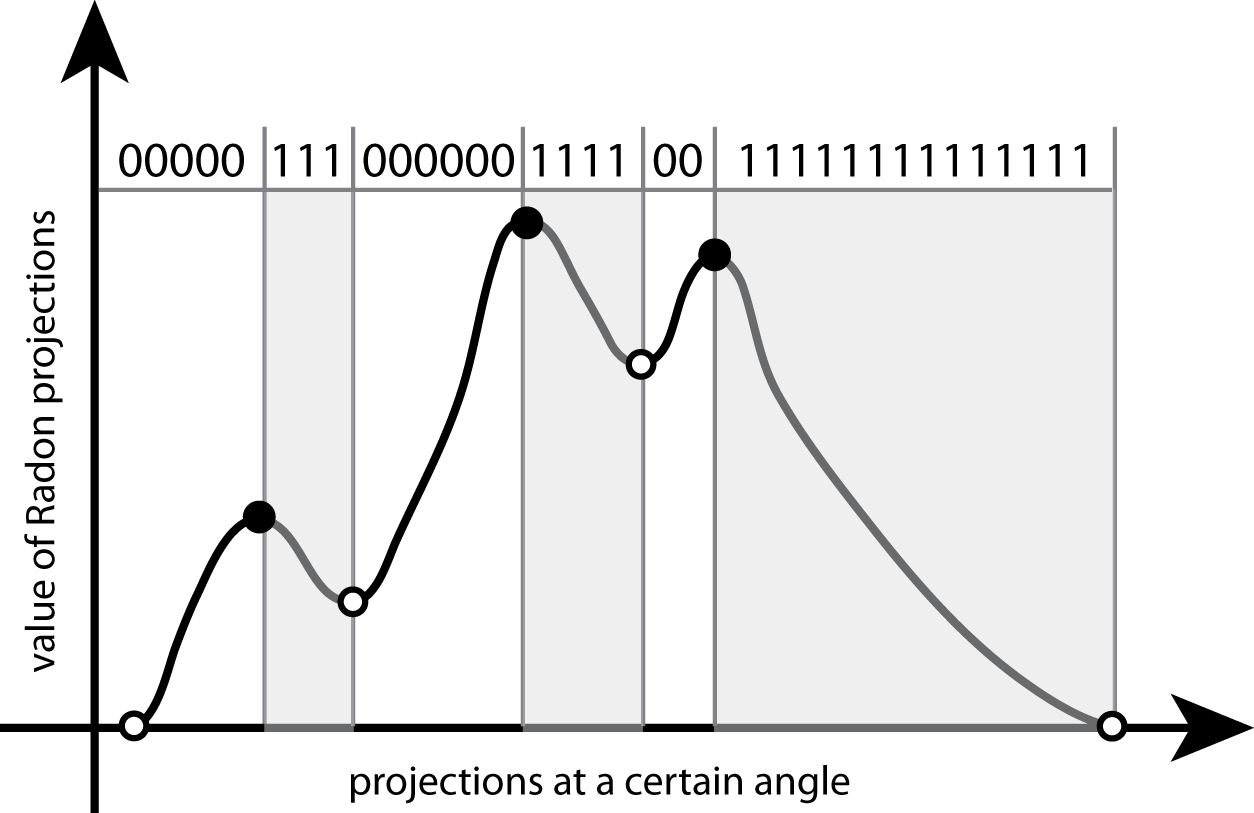}
\caption{MinMax Radon Barcodes. Projections at a certain angle are smoothed to find minimums and maximums. All bins between a minimum and a maximum are assigned 0, whereas all bins between a maximum and a minimum are assigned 1. }
\label{fig:minmaxRBC}
\end{center}
\end{figure}

Figure \ref{fig:minmaxRBC} illustrates how MinMax Radon barcodes are generated for a given angle $\theta$. The order of assignments for zeros/ones for transitions from min/max to max/min, of course, is just a convention and hence must be maintained consistently within a given application. 

\begin{figure*}[htb]
\begin{center}
\includegraphics[width=0.20\columnwidth,height=0.20\columnwidth]{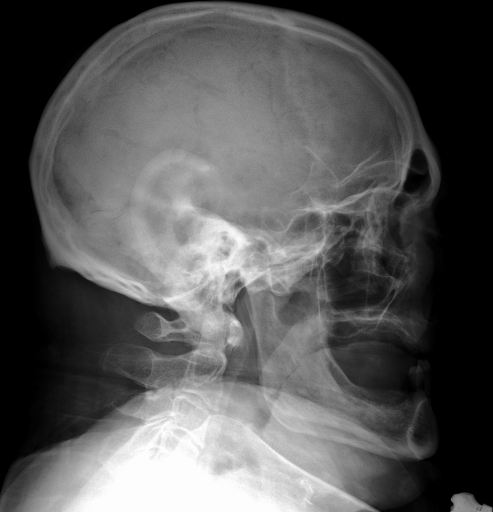} \quad
\includegraphics[width=0.20\columnwidth,height=0.20\columnwidth]{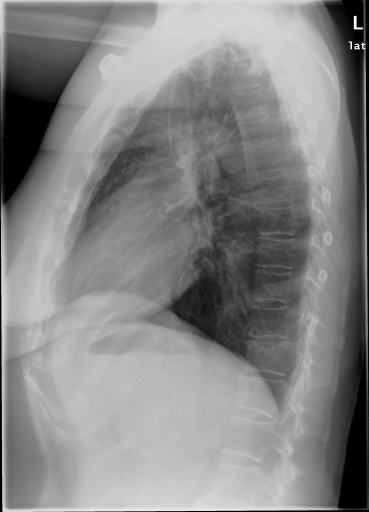} \quad
\includegraphics[width=0.20\columnwidth,height=0.20\columnwidth]{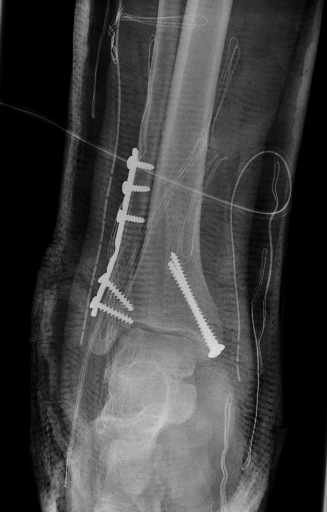} \quad 
\includegraphics[width=0.20\columnwidth,height=0.20\columnwidth]{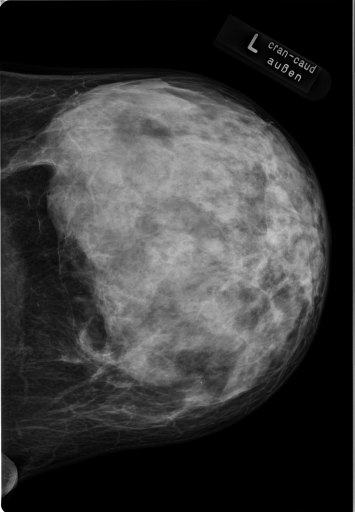} \\  \vspace{0.05in}
\includegraphics[width=0.20\columnwidth, height=0.6cm]{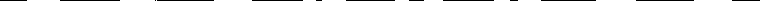} \quad
\includegraphics[width=0.20\columnwidth, height=0.6cm]{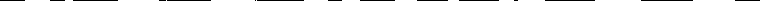} \quad
\includegraphics[width=0.20\columnwidth, height=0.6cm]{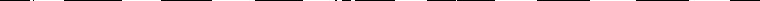} \quad 
\includegraphics[width=0.20\columnwidth, height=0.6cm]{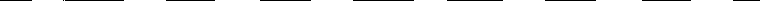} \\ \vspace{0.05in}
\includegraphics[width=0.20\columnwidth, height=0.6cm]{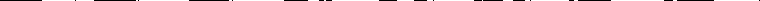} \quad
\includegraphics[width=0.20\columnwidth, height=0.6cm]{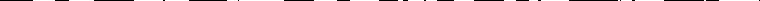} \quad
\includegraphics[width=0.20\columnwidth, height=0.6cm]{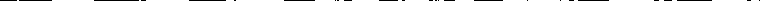} \quad
\includegraphics[width=0.20\columnwidth, height=0.6cm]{S3type4.png}
\caption{Local Radon barcodes (top barcodes) and MinMax Radon barcodes (bottom barcodes) for four sample images from IRMA dataset. Images were resized to $64\!\times\!64$ and projected at 8 angles.}
\label{fig:barcodeExamples}
\end{center}
\end{figure*}

Figure \ref{fig:barcodeExamples} shows barcodes for three images from IRMA dataset. For each image, both barcodes are provided to examine the visual difference between Radon barcodes using local thresholding and MinMax Radon barcodes as introduced in this paper. The former appears to be a coarse encoding as the latter shows finer bit distribution.

\section{Experiments}
\label{sec:exp}
In this section, we first describe the IRMA dataset, the benchmark data we used. The error calculation is reviewed next. Subsequently, we report two series of experiments to validate the performance of the proposed MinMax Radon barcodes for medical image retrieval. The first series of experiments compares MinMax barcodes against the recently introduced Radon barcodes using local thresholding using $k$-NN search. The second series of experiments compare MinMax barcodes against SURF and BRISK when hashing is used for matching.  

\subsection{Image Test Data}
\label{subsec:IRMA}
The Image Retrieval in Medical Applications (IRMA) database \footnote{http://irma-project.org/} is a collection of more than 14,000 x-ray images (radiographs) randomly collected from daily routine work at the Department of Diagnostic Radiology of the RWTH Aachen University \footnote{http://www.rad.rwth-aachen.de/} \cite{Lehmann2003},\cite{Mueller2010}. All images are classified into 193 categories (classes) and annotated with the ``IRMA code'' which relies on class-subclass relations to avoid ambiguities in textual classification \cite{Mueller2010},\cite{Lehmann2006}. The IRMA code consists of four mono-hierarchical axes with three to four digits each: the technical code T (imaging modality), the directional code D (body orientations), the anatomical code A (the body region), and the biological code B (the biological system examined). The complete IRMA code subsequently exhibits a string of 13 characters, each in $\{0,\dots,9;a,\dots,z\}$:

\begin{equation}
\textrm{TTTT-DDD-AAA-BBB}. 
\end{equation}

\begin{figure*}[htb]
\centering 
\subfigure[\tiny 1121-127-700-500]{\label{fig:a}\includegraphics[width=27mm,height=27mm]{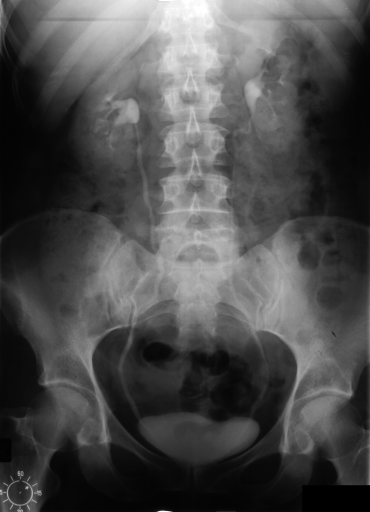}}
\subfigure[\tiny1121-120-942-700]{\label{fig:b}\includegraphics[width=27mm,height=27mm]{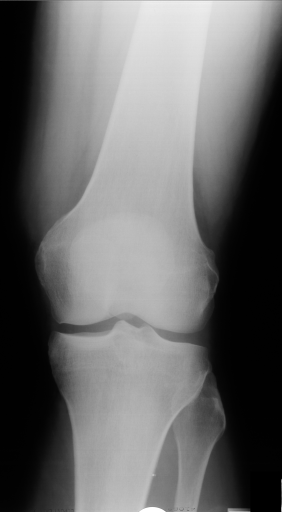}}
\subfigure[\tiny1123-127-500-000]{\label{fig:c}\includegraphics[width=27mm,height=27mm]{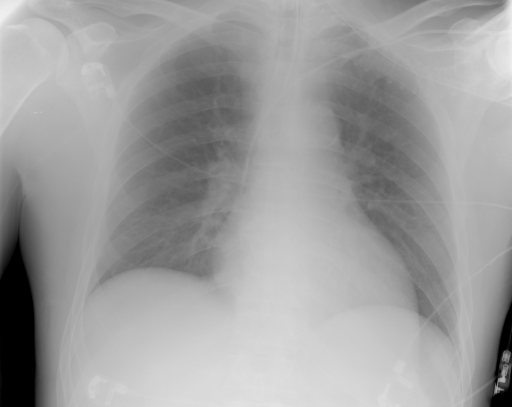}} \\
\subfigure[\tiny1121-120-200-700]{\label{fig:d}\includegraphics[width=27mm,height=27mm]{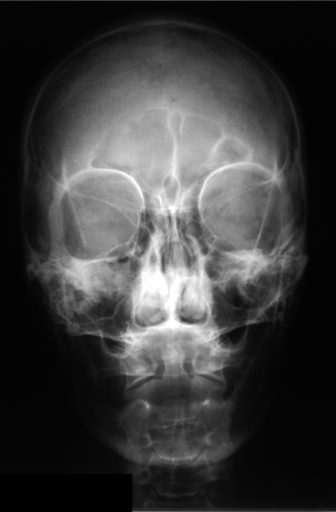}}
\subfigure[\tiny1121-120-918-700]{\label{fig:e}\includegraphics[width=27mm,height=27mm]{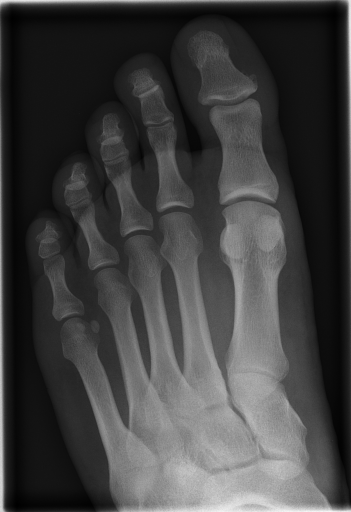}}
\subfigure[\tiny1121-220-310-700]{\label{fig:f}\includegraphics[width=27mm,height=27mm]{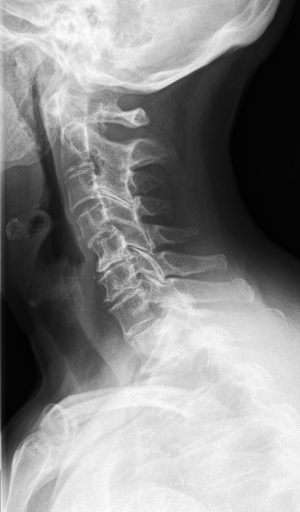}}
\caption{Sample x-ray images with their IRMA codes TTTT-DDD-AAA-BBB.}
\label{fig:IRMASamples}
\end{figure*}

Details of the IRMA database is described in literature \cite{Lehmann2003},\cite{Lehmann2006},\cite{Mueller2010}. IRMA dataset offers 12,677 images for training and 1,733 images for testing. Figure \ref{fig:IRMASamples} shows some sample images from the dataset long with their IRMA code in the format TTTT-DDD-AAA-BBB.

\subsection{Error Calculation}
\label{sec:error}
We used the formula provided by \emph{ImageCLEFmed09} to compute the error between the IRMA codes of the testing images (1,733 images) and the first hit retrieved from all indexed images (12,677 images) in order to evaluate the performance of the retrieval process. We then summed up the error for all testing images. The formula is provided as follows:

\begin{equation} \label{equation:IRMA1}
E_{Total} = \sum_{m=1}^{1733} \sum_{j=1}^{4} \sum_{i=1}^{l_{j}} \frac{1} {b_{l_{j},i}} \frac {1} {i} \delta (I_{l_{j},i}^{m}, \tilde{I}_{l_{j},i}^{m})
\end{equation}

Here, $m$ is an indicator to each image, $j$ is an indicator to the structure of an IRMA code, and $l_{j}$ refers to the number of characters in each structure of an IRMA code. For example, consider the IRMA code: 1121-4a0-914-700, $l_{1}=4$, $l_{2}=3$, $l_{3}=3$ and $l_{4}=3$. Here, $i$ is an indicator to a character in a particular structure. Here, $l_{2,2}$ refers to the character ``a'' and $l_{4,1}$ refers to the character ``7''. $b_{l_{j},i}$ refers to the number of branches, i.e. number of possible characters, at the position $i$ in the $l_{j}^{th}$ structure in an IRMA code. $I^{m}$ refers to the $m^{th}$ testing image and $\tilde{I}^{m}$ refers to its top 1 retrieved image. $\delta (I_{l_{j},i}^{m}, \tilde{I}_{l_{j},i}^{m})$ compares a particular position in the IRMA code of the testing image and the retrieved image. It then outputs a value in \{0, 1\} according to the following rules:

\begin{equation} \label{equation:IRMA2}
\delta (I_{l_{j},i}^{m}, \tilde{I}_{l_{j},i}^{m})=
\begin{cases}
0, & I_{l_{j},h}^{m} = \tilde{I}_{l_{j},h}^{m} \forall h \leq i \\
1, & I_{l_{j},h}^{m} \neq \tilde{I}_{l_{j},h}^{m} \exists h \leq i
\end{cases}
\end{equation}

We used the Python implementation of the above formula provided by ImageCLEFmed09 to compute the errors \footnote{http://www.imageclef.org/}. 

\subsection{Results}
We report two series of experiments in this section: First we compare the proposed MinMax Radon barcodes with the local thresholding barcodes to validate their retrieval performance. Second, we compare MinMax Radon barcodes with SURF (with non-binary features) and BRISK (with binary features). All experiments were conducted using IRMA x-ray images. 

\subsection{MinMax versus Thresholding}
We applied both types of Radon barcodes on IRMA dataset. We first used 12,677 images and indexed them with both types of barcodes. Then, we used 1,733 remaining images to measure the retrieval error of each barcode type according to IRMA code error calculation (see section \ref{sec:error}). To measure the similarity between two given barcodes we used Hamming distance. For conducting the actual search, we used $k$-NN with $k=1$ (no pre-classification was used). Table \ref{tab:minmaxresults} shows the results. The retrieval error clearly drops when we use MinMax barcodes. The reduction for 8 or 16 projection angles is around 15\%.

\setlength{\tabcolsep}{4pt}
\begin{table}[t]
\begin{center}
\caption{Comparing MinMax barcodes with thresholding barcodes as described in \cite{Tizhoosh2015}. Images were normalized into $32\!\times\!32$. Projections angles were equi-distance in $[0^\circ,180^\circ)$. A total of 12,677 images were indexed. Retrievals were run for  1,733 unseen images. }
\label{tab:minmaxresults}
\begin{tabular}{lll}
\hline\noalign{\smallskip}
& 8 angles & 16 angles \\ 
\noalign{\smallskip}
\hline
\noalign{\smallskip}
Thresholding Barcodes & 605.83 & 576.45 \\ 
MinMax Barcodes & 509.24 & 489.35 \\ 
Error reduction & 15.94\% & 15.11\% \\
\hline
\end{tabular}
\end{center}
\end{table}
\setlength{\tabcolsep}{1.4pt}

\subsection{Barcodes versus SURF and BRISK}
In this series of experiments, we also examined SURF (as a non-binary method) and BRISK (as a binary method). To our knowledge, this is the first time that these methods are being applied on IRMA images. Using $k$-NN as before was not an option because initial experiments took considerable time as SURF and BRISK appear to be slower than barcodes. Hence, we used locality-sensitive hashing (LSH) \cite{Indyk1998}to hash the features/codes into patches of the search space that may contain similar images\footnote{Matlab code: http://goo.gl/vFYvVJ}. We made several tests in order to find a good configuration for each method. As well, the configuration of LSH (number of tables and key size for encoding) was subject to some trial and errors. We set the number of tables for LSH to 30 (with comparable results for 40) and the key size to a third of the feature vectors' length. We selected the top 10 results of LSH and chose the top hit based on highest correlation with the input image for each method. The results are reported in Table \ref{tab:SURFBRISK}.

As apparent from the results, not only do SURF and BRISK deliver higher error rates than MinMax barcodes, but also for many cases, they fail to provide any features at all. Hence, we measured their error only for the cases they successfully located key points and extracted features. For failed cases we just incremented the number of failures. 

\setlength{\tabcolsep}{4pt}
\begin{table}[t]
\begin{center}
\caption{Comparing MinMax barcodes with SURF and BRISK. Images were normalized into $32\!\times\!32$ (12,677 indexed images and 1,733 test images). LSH was used for the actual search to deliver 10 matches. The top hit was found via correlation measurement with the query image. }
\label{tab:SURFBRISK}
\begin{tabular}{llll}
\hline\noalign{\smallskip}
Method & Total Error & Failure & $\bar{t}$ (s) \\ \hline
SURF & 526.05 & 4.56\% & 6.345\\
BRISK & 761.96 & 1.095\% & 6.805\\
MinMax RBC & 415.75 & 0.00\% & 0.537 \\ 
\hline
\end{tabular}
\end{center}
\end{table}
\setlength{\tabcolsep}{1.4pt}

\section{Summary and Conclusions}
In this paper, we improved Radon barcodes by introducing a new encoding scheme called MinMax Radon barcodes. Instead of local thresholding we encode the projection values for each angle of Radon transform by examining the extreme values of the projection curvature. We employed IRMA dataset with 14,410 x-ray images to validate the proposed MinMax Radon barcodes. The results confirm 15\% reduction in retrieval error for IRMA images. 

We also compared the proposed MinMax Radon barcodes with SURF and BRISK. Using locality-sensitive hashing (LSH), we applied SURF and BRISK, for the first time, on IRMA images. We found that MinMax Radon barcodes are both more accurate (lower error), more reliable (no failure) and faster (shorter average time $\bar{t}$) compared with SURF and BRISK for this dataset.

Radon barcodes seem to have a great potential for medical image retrieval. One question that needs to be answered is which projection angles may provide more discrimination in order to make Radon barcodes even more accurate. Other schemes for encoding Radon projections may need to be investigated as well.

\bibliographystyle{splncs}
\bibliography{isvc_submission}

\end{document}